\pdfoutput=1

\documentclass[11pt, table]{article}

\usepackage[]{EMNLP2022}
\usepackage{multirow}
\usepackage{times}
\usepackage{booktabs}
\usepackage{latexsym}
\usepackage{supertabular}
\usepackage{csquotes}
\usepackage{xcolor}
\PassOptionsToPackage{table,xcdraw}{xcolor} 
\usepackage{xcolor}
\usepackage{amsfonts}
\usepackage{float}
\usepackage[graphicx]{realboxes}
\usepackage{graphicx}
\usepackage{amsmath}
\usepackage{booktabs}
\usepackage{subfig}
\usepackage{multirow}
\usepackage{siunitx}

\usepackage[T1]{fontenc}

\usepackage[utf8]{inputenc}

\usepackage{microtype}

\usepackage{inconsolata}
\usepackage{xspace}

\title{Optimizing text representations to capture \\ (dis)similarity
between political parties}

\author{Tanise Ceron$^\triangle$ \qquad Nico Blokker$^\Box$ \qquad Sebastian Pad\'o$^\triangle$ \\
  $^\triangle$ Institute for Natural Language Processing, University of Stuttgart, Germany  \\
    $^\Box$ Research Center on Inequality and Social Policy, University of Bremen, Germany \\
  \texttt{\{tanise.ceron,pado\}@ims.uni-stuttgart.de}, \texttt{blokker@uni-bremen.de} \\
}

\begin{document}
\maketitle
\begin{abstract}
Even though fine-tuned neural language models have been pivotal in
enabling “deep” automatic text analysis, optimizing text
representations for specific applications remains a crucial
bottleneck.  In this study, we look at this problem in the context of
a task from computational social science, namely modeling pairwise
similarities between political parties. Our research question is what
level of structural information is necessary to create robust text
representation, contrasting a strongly informed approach (which uses
both claim span and claim category annotations) with approaches that
forgo one or both types of annotation with document structure-based
heuristics. Evaluating our models on the manifestos of German parties
for the 2021 federal election. We find that heuristics that maximize
within-party over between-party similarity along with a normalization
step lead to reliable party similarity prediction, without the need
for manual annotation.
\end{abstract}


\section{Introduction}
\label{sec:intro}

A party manifesto, also known as electoral program, is a document in
which parties express their views, intentions and motives for the next
coming years. Since this genre of text is written not just to inform,
but to persuade potential voters that the parties compete for
\citep{budge2001}, it provides a strong basis to understand the
position taken by parties according to various policies because of its
direct access to the parties' opinions. Political scientists study the
contents of party manifestos, for instance, to investigate parties'
similarity with respect to the several policies
\citep{budge2003validating}, to predict party coalitions
\citep{druckman2005influence}, and to evaluate the extent to which the
parties that they vote for actually corresponds to their own world
view \citep{mcgregor2013measuring}.

To carry out systematic analyses of party relations while taking into
account differences in style and level of detail, these analyses are
increasingly grounded in two types of manual annotation about
\textit{claims}, statements that contain a position or a view towards
an issue, that can be argued or demanded for
\citep{koopmans1999political}: First, \textit{abstract claim
categories} \citep{manifesto} are used to group together diverse
forms and formulations of demands. Second, annotation often includes
the \textit{stance} that parties take towards specific political
claims to abstract away from the many ways to express support or
rejection in language. In addition, these types of annotation offer a
direct way to empirically ground party similarity in claims and link
these to concrete textual statements. At the same time, such manual
annotation is extremely expensive in terms of time and resources and
has to be repeated for every country and every new election.


In this paper, we investigate the extent to which this manual effort
can be reduced given appropriate text representations. We build on the advances made in recent years in neural language models for text representations and present a series of fine-tuning designs based on manifesto texts to compute party similarities. Our main hypothesis is that the proximity between groups can be more easily captured when the model receives adequate indication of the differences between groups (and their stances) and this can be done via fine-tuning for instance. This can be achieved by using signal that is freely available in the manifestos' \textit{document structure}, such as groupings by party or topic. Information of this  type can serve as an alternative feedback for fine-tuning in order to create robust text representations for analysing party proximity.  


We ask three specific questions: (1) How to create robust representations for identifying the similarity between groups such as in the case of party relations? (2) What level of document structure is necessary for this purpose?  (3) Can computational methods capture the relation between parties in unstructured text? We empirically investigate these questions on electoral programs from the German 2021 elections, comparing party similarities against a ground truth built from structured data. We find that our hypothesis is borne out: We can achieve competitive results in modelling the party proximity with textual data provided that the text representations are optimized to capture the differences across parties and normalized to fall in a certain distribution that is appropriate for computing text similarity. More surprisingly, we find that completely unstructured data reach higher correlations than more informed settings that consider exclusively claims and/or their policy domain. We make our code and data available for replicability.\footnote{\url{https://github.com/tceron/capture_similarity_between_political_parties.git}}


\paragraph{Paper structure.} The paper is structured as
follows. Section~\ref{sec:related-work} provides an overview of
related work. Section~\ref{sec:datasets} describes the data we work
with and our ground truth. Section~\ref{sec:methods} presents our
modeling approach. Sections~\ref{sec:experimental-setup}
and~\ref{sec:results-discussion} discuss the experimental setup and
our results. Section~\ref{sec:conclusion} concludes.

\section{Related Work}
\label{sec:related-work}

\subsection{Party Characterization}

The characterization of parties is an important topic in political
science, and has previously been attempted with NLP models. Most
studies, however, have focused on methods to place parties along the
left to right ideological dimension. For instance, an early example is
\citet{laver2003extracting} who investigate the scaling of political
texts associated with parties (such as manifestos or legislative
speeches) with a bag of words approach in a supervised fashion, with
position scores provided by human domain experts. Others, instead,
have implemented unsupervised methods for party positioning in order
to avoid picking up on biases of the annotated data and to scale up to
large amounts of texts from different political contexts while still
implementing word frequency methods \citep{slapin2008scaling}. More
recent studies have sought to overcome the drawbacks of word frequency
models such as topic reliance and lack of similarity between
synonymous pairs of words, e.g. \citet{glavas-etal-2017-unsupervised}
and \citet{nanni2022political} implement a combination of
distributional semantics methods and a graph-based score propagation
algorithm for capturing the party positions in the left-right
dimension.

Our study differs from previous ones in two main aspects. First, our
aim is not to place parties a left-to-right political dimension but to
assess party similarity in a latent multidimensional space of policy
positions and ideologies. Second, our focus is not on the use of
specific vocabulary, but on representations of whole sentences. In
other words, our proposed models work well if they manage to learn how
political viewpoints are expressed at the sentence level in party
manifestos.

\begin{table*}[]
\centering
\begin{tabular}{lp{0.7\textwidth}p{0.15\textwidth}}
\toprule
\textbf{Party}     & \textbf{Sentence}                                                                                                                                                                                                                                                               & \textbf{Domain}                                                                                                                                    \\ \midrule
AfD       & People's insecurities and fears, especially in rural regions, must be taken seriously.                                                                                                                                      & Social Groups                                   \\ 
CDU       & We want to strengthen our Europe together with the citizens for the challenges of the future.                                                                                                                         & External \par Relations        \\

 Linke & The policies of federal governments that ensure private corporations and investors can make big money off our insurance premiums, co-pays and exploitation of health care workers are endangering our health! & Political \par System                     \\ 
FDP       & In this way, we want to create incentives for a more balanced division of family work between the parents. & Welfare and Quality of Life \\ 
Grüne    & After the pandemic, we do not want a return to unlimited growth in air traffic, but rather to align it with the goal of climate neutrality.                                                                   & Economy                          \\ 
SPD       & We advocate EU-wide ratification of the Council of Europe's Istanbul Convention as a binding legal norm against violence against women.                                                                 & Fabric of \par Society     \\
\bottomrule
\end{tabular}
\caption{Examples from the 2021 party manifestos and their annotated domains. }
\label{tab:ex-manifesto21}
\end{table*}

\subsection{Optimizing Text Representations for Similarity}

\paragraph{Fine Tuning.}
Recent years have seen rapid advances in the area of neural language
models, including models such as BERT, RoBERTa or GPT-3
\citep{devlin-etal-2019-bert,
liu2020roberta,NEURIPS2020_1457c0d6}. The sentence-encoding
capabilities of these models make them generally applicable to text
classification and similarity tasks \citep{cer-etal-2018-universal}.
Both for classification and for similarity, it was found that
pre-trained models already show respectable performance, but
fine-tuning them on task-related data is crucial to optimize the
models' predictions -- essentially telling the model which aspects of
the input matter for the task at hand.

On the similarity side, a well-known language model is Sentence-BERT
\citet{reimers2019sentencebert}, a siamese and triplet network based
on BERT \citep{devlin-etal-2019-bert} or RoBERTa
\citep{liu2020roberta} which aims at better encoding the similarities between sequences of text. Sentence-BERT (SBERT) comes with
its own fine-tuning schema which is informed by ranked pairs or
triplets and tunes the text representations to respect the preferences
expressed by the fine-tuning data. Of course, this raises the question
of how to obtain such fine-tuning data: The study experiments both
with manually annotated datasets (for entailment and paraphrasing
tasks) and with the use of heuristic document structure information,
assuming that sentences from the same Wikipedia section are
semantically closer and sentences from different sections are
further away. Parallel results are also found by
\citet{gao-etal-2021-simcse} in their SimCSE model, which reach even
better results when fine-tuning with contrastive learning: They also
compare a setting based on manually annotated data from an inference
dataset with a heuristic setting based on combining a pair of sentences with its drop-out version as positive examples and different pairs as negative examples. 

Both studies find slightly lower performance for the heuristic
versions of their fine-tuning datasets, but still obtain a relevant
improvement over the non-fine-tuned versions of their models, pointing
to the usefulness of heuristically generated fine-tuning data, for
example based on document structure.

\paragraph{Postprocessing to Improve Embeddings}
A problem of the use of neural language models to create text
representations that was recognized recently concerns the
distributions of the resulting embeddings: They turn out to be highly
anisotropic \citep{ethayarajh-2019-contextual, gao-2019}, meaning that
their semantic space takes a cone rather than a sphere format - in the
former two random vectors are highly correlated while in the latter
they should be highly uncorrelated. This can cause similarities
between tokens or sentences to be very similar even when they should
not. To counteract this tendency, \citet{li-etal-2020-sentence} impose
an isotropic distribution onto the embeddings via a flow-based
generative model. \citet{Su2021WhiteningSR} propose a lightweight,
even slightly more effective approach: The text embeddings undergo a
linear so-called whitening transformation, which ensures that the
bases of the space are uncorrelated and each have a variance
of~1.

\section{Data}
\label{sec:datasets}

Before we describe the methods we will use, we describe our textual
basis and the ground truth we will aim to approximate.

\subsection{The Manifesto Dataset}

As stated above, we are interested in deriving party representations
from party manifestos.  Party manifestos generally contain sections
roughly separated by policy topics, however, some party manifestos are
organized more strictly by topics than others. For this reason, we
utilize the manifesto dataset provided by the Manifesto Project
\citep{manifesto}, which provides manifestos from around the world and
offers consistent markup of policy domains and categories 
\footnote{More information on
  \url{https://manifesto-project.wzb.eu/information/documents/corpus}}.

More specifically, every sentence from the manifestos is annotated
with domain names and categories. In this paper, consistent with our
goal of reducing annotation effort, we consider only the domain. The
domain corresponds to a broad policy field such as \enquote*{political
  system} and \enquote*{freedom and democracy}. In most cases, an
entire sentence is annotated with a single domain, but some sentences
have been split when falling into two distinct domains. Nearly every
sentence is annotated with a domain label, except the introduction and
end sections which usually contain an appeal to the voter and do not
belong to any policy category.

For reasons that will become clear in the next subsection, we focus on
German data and use the party manifestos written by the six main
German parties (CDU/CSU, SPD, Grüne, Linke, FDP, AFD) for the federal
elections in 2013, 2017 and 2021.  Table \ref{tab:ex-manifesto21}
shows some examples of sentences with their respective domain
names. Due to space constraints, more information about the description of
the dataset is found in appendix \ref{sec:appendix}.

\subsection{Ground Truth: Wahl-o-Mat}
\label{sec:ground-truth:-wahl}

A problem with the task of predicting party proximity is to find a
suitable ground truth against which to evaluate the models. In this
study, we make use of a highly structured dataset, Wahl-o-Mat (WoM)
from which we can construct a ground truth of party similarities with
minimal manual involvement.

Wahl-o-Mat (WoM, \citet{wagner2012}) is an online application that
provides voting advice. The application collects users' stances on a
range of policy issues via a questionnaire. There are 38 issues in
total and they cover a wide range of topics, e.g. \enquote*{Germany
should increase its defense spending} or \enquote*{The promotion of
wind energy is to be terminated}.  The users' stances are then matched
against those of the German parties in order to suggest the closest
choices for users. The database behind WoM consists of the stances
that each party takes towards each policy issue, which can be
\enquote*{agree}, \enquote*{disagree}, or \enquote*{neutral}.

WoM provides each user with a ``percentage overlap'' that they have
with the different parties, suggesting that the set of policy issues
and the stances are an informative basis for computing positional
similarity \citep{wagner2012}. In this spirit, we define as our ground
truth the \textit{party distance matrix} which we obtain by
representing each party by its vector of stances (represented -1, 1,
0) towards the different policy issues and computing the Hamming (L1)
distances among them. Such distance calculations are used by
political scientists to understand the overall (dis)similarity between party and voters \citep{mcgregor2013measuring}.

\begin{figure}[tb]%
    \centering
    \subfloat[\centering Distances between parties]{{\includegraphics[width=\columnwidth]{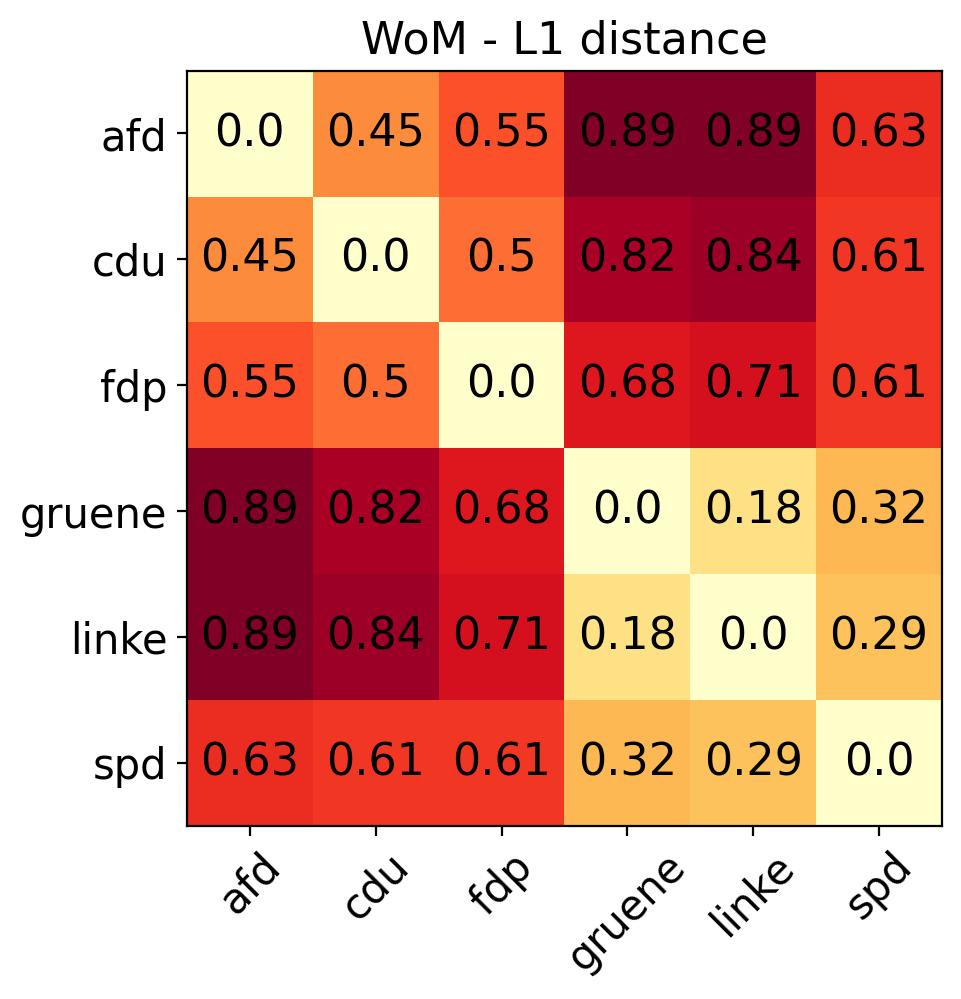} }}%
    \qquad
    \subfloat[\centering Aggloremative clustering]{{\includegraphics[width=\columnwidth]{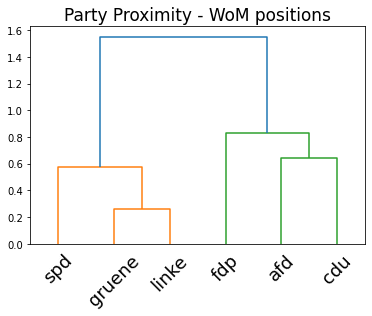} }}%
    \caption{Based on Wahl-o-Mat policy positions.}%
    \label{fig:wom}%
\end{figure}

Figure \ref{fig:wom}a shows the distance matrix between parties: the
higher the distance, the more they disagree on WoM policy issues.
Figure \ref{fig:wom}b visualizes the ground truth differently, as an
agglomerative clustering of the distance matrix. This ground truth
arguably stands up to scrutiny: The two most left-oriented parties,
Grüne (greens) and Linke (left), are most similar (distance 0.18), due
to their similar environmental programs and shared concern about
foreign policy. They are then most similar to social democratic
SPD. On the other main branch of the clustering tree, which covers the
right-oriented parties, AFD (right wing) and CDU/CSU (center
conservative) are most similar, although less than the left parties
(distance 0.45). Finally, the liberal party FDP groups with the
conservative parties, but reluctantly so: it
assumes a kind of bridge position between the left and right oriented
parties.

\section{Methods}
\label{sec:methods}

We describe our method in three steps: (a) we define a set of
informative text representations models; (b) we compute party similarities,
parallel to Section~\ref{sec:ground-truth:-wahl}, on the basis of
these text representations; (c) we post-process the data.

\subsection{Building Informative Text Representations}
\label{subsec:text-rep}

The first step is to build text representations that are informative
for party similarity. As sketched above, we use neural language models
(NLMs) as the current state of the art. This involves selecting a base
embedding model and defining the different fine-tuning schemes.

\paragraph{Base embedding model: SBERT.} We choose SBERT as the basis
for our models. With its focus on sentence similarity and its
computational efficiency, it is arguably the most appropriate model
for our goals. Pre-trained SBERT without any fine-tuning  \footnote{Pre-trained model:
  \url{https://huggingface.co/sentence-transformers/paraphrase-multilingual-mpnet-base-v2}} serves
directly as our first model.

\paragraph{Fine-tuning SBERT.}
Fine-tuning of SBERT can take place in different ways, but given our
type of data, we use the triplet objective function where the model
receives as input an anchor sentence \emph{a}, a positive sentence
\emph{p} that is similar to the anchor sentence and a negative
sentence \emph{n} unrelated to both previous sentences. The objective
of the fine-tuning is to minimize
\begin{equation}
    max(\left\Vert S_a - S_p \right\Vert - \left\Vert S_a - S_n \right\Vert + \epsilon, 0)
\end{equation}
which encourages the model to learn that $S_p$ is at least $\epsilon$
closer to $S_a$ than to $S_n$. $\left\Vert \cdot \right\Vert$ is the
distance metric, which is kept as the default Euclidean \footnote{Loss
  function and more details on:
  \url{https://www.sbert.net/docs/package_reference/losses.html\#sentence_transformers.losses.BatchAllTripletLoss}}.
We experiment with two ways of constructing triplets for fine-tuning,
first by \emph{domain} and then by \emph{party}.

\paragraph{SBERT$_{domain}$} follows the same logic as in
\citet{ein-dor-etal-2018-learning} with the Wikipedia sections (and
replicated in \citet{reimers2019sentencebert}). We use the domain
information from the manifestos (cf. Section~\ref{sec:datasets}) to
construct triplets: The anchor and the positive sentences are part of
the same domain and the negative sentence is from a different domain
across party manifestos. The hypothesis is that aligning sentences by
topic should help the model focus on relevant policy distinctions
across parties.

\paragraph{SBERT$_{party}$}, in contrast, intends to learn the
distinction between the way parties express their claims or their
ideologies and opinion. Here, we construct triplets by combining
anchor sentences with positive sentences from the same party --
irrespective of the domain -- and negative sentences from the other
parties' manifestos. The hypothesis of this setup is that the
embeddings incorporate the parties' stances along with the way that
particular sentences are presented, or styles used. We assume that
many aspects of the text contribute to capturing the stance such as
sentiment, text style and word usage.


\subsection{Four Models for Party Similarities}
\label{sec:four-models-compute}


\newcommand{\cldom}{\textsc{ClaimDom}\xspace}
\newcommand{\cl}{\textsc{Claim}\xspace}
\newcommand{\dom}{\textsc{Dom}\xspace}
\newcommand{\none}{\textsc{None}\xspace}

\begin{table}[tb]
\centering
\setlength{\tabcolsep}{4pt}
\begin{tabular}{llll}
\textbf{ID} & \textbf{Grouping} & \textbf{Filtering}  & \textbf{Infor.} \\ \hline
\cldom              & Domain        & Claims only & ++++               \\ \hline
\cl                  & -                 & Claims only         & +++                \\ \hline
\dom                 & Domain        & All sentences            & ++                \\ \hline
\none                 & -                 & All sentences             &  +               \\ 
\end{tabular}%
\caption{Models for the computational of party similarity, varying in the amount of information used}
\label{tab:eval-desc}
\end{table}

With the methods described in the previous subsection, we can obtain
representations for individual sentences. We now need to define how to
\textit{aggregate} these sentences into global party
representations -- or rather, their similarities.

Table \ref{tab:eval-desc} shows four aggregating strategies that
differ in the amount of information that they take into account. They
differ in two main dimensions: (a), the \textit{grouping}: is the
similarity computed globally, over the complete manifestos, or domain
by domain  (b), the \textit{filtering}: is the similarity based on
all sentences in the manifestos, or only on sentences that contain
concrete claims (cf. Section~\ref{sec:intro}).

Regarding grouping, we hypothesize that it is easier for language
models to assess the proximity between parties if sentences from
matching topics are compared. Similarly, we expect that filtering by
claims serves to focus the models on the \enquote*{core} of the
parties' policies.

\newcommand{\mysim}{\mathop{\mathrm{sim}}}
\newcommand{\mytsim}{\mathop{\mathrm{tsim}}}

\paragraph{\cldom: using claims and domains.} In this, the most
informed, model, we represent parties by the claims that they make,
compare these claims by domain, and then average the by-domain
similarities. Formally, let $\vec s$ be the embedding produced for a sentence by an (implicit) encoder model, $cl(T)$ the set of claim sentences
contained a text T, and $dom(P,i)$ the set of sentences for domain $i$ in the manifesto of a party $P$. Then we can define the representation of a domain (Equation 1), the similarity for domain $i$ (Equation 2), and a
global similarity (Equation 3):
  \begin{align}
    \vec{dom}(P,i) & = \sum_{s \in cl(dom(P,i))} \vec s \\
  \mysim(P_1,P_2,i) & = \cos(\vec{dom}(P_1,i), \\ 
          & \phantom{ = \cos( }\, \vec{dom}(P_2,i)) \notag \\
    \mysim(P_1, P_2) & = \frac{1}{|Dom|} \sum_{i} \mysim(P_1,P_2,i)    
\end{align}%

\begin{figure}[tb]
  \centering
    \includegraphics[width=0.8\columnwidth]{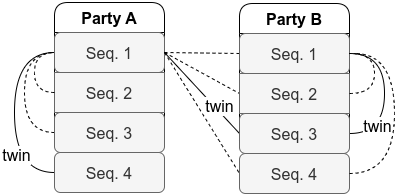}
  \caption{Twin matching: Solid lines mean pairings of maximal similarity.}
  \label{fig:cos-simb}
\end{figure}

\paragraph{\cl: using claims, but no domains.} To compute similarities
without domain information, we could simply average over all sentences
of the manifestos. However, pilot experiments showed that this
procedure resulted in a severe loss of information. To avoid this, we
introduce a method called \textit{twin matching}, visualized in
Figure~\ref{fig:cos-simb}. Twin matching maps each sentence in one
manifesto to its nearest neighbor in the other manifesto (Equation 5) -- in
most cases, this will be a sentence of the same domain. Furthermore,
we normalize the similarity to the twin by dividing by the
maximum inter-claim similarity to both manifestos, and average over all 
sentences in the manifesto (Equation 7). Our hypothesis is that this procedure provides
an approximating to domain-based grouping without the need for
explicit domain labeling.

Formally, let $tw(s,T)$ denote the nearest neighbor, or twin, of sentence $s$ in text $T$:
\begin{align}
    tw(s,T) & = \arg\max_{t \in T} \cos(s,t)
\end{align}
Then the maximum inter-claim similarity $C$ of a manifesto $P$, is
\begin{align}
    C(P) = \max_{p, p' \in cl(P) \wedge p \not= p'} cos(p,p')
\end{align}

Then the similarity of two texts is:
  \begin{align}
& \mysim(P_1, P_2) = \\
& \quad \sum_{s \in cl(P_1)} \frac{\cos(s, tw(s, P_2)) }{|cl(P_1)| (C(P_1)+C(P_2))} \notag
\end{align}%

\paragraph{\dom: using domains, but no claims.} This model is identical
to \cldom, but uses all sentences instead of just claims in Equation~(2).

\paragraph{\none: using neither domains nor claims.} This model is
identical to \cl, but uses all sentences instead of just claims in
Equations~(6) and~(7).

\subsection{Post-processing}


As mentioned in Section \ref{sec:related-work}, sequence
representations should form an isotropic space for good similarity
prediction. Therefore, we also experiment with post-processed
embeddings of the sentences by applying whitening transformation to our embeddings as suggested in \citet{Su2021WhiteningSR}. Following their
normalization procedure, we start with a matrix $\mathbb{R}^{n\times d}$
representing \emph{n} sequence vectors from a given encoding model
with dimension \emph{d}.\footnote{The pre-trained model we use has 768 dimensions.} Then, matrix W
($\mathbb{R}^{d\times d}$) is computed through singular value
decomposition (SVD) and saved along with the mean vector $\mu$
($\mathbb{R}^{1\times d}$) retrieved from the initial input embedding
matrix. Finally, every vector ($\tilde{x}_i$) of interest for the analysis is converted into our final representation as in $\tilde{x}_i = (x_i - \mu)W$. 

\citet{Su2021WhiteningSR} compute W and $\mu$ either with the  data
from the task at hand (train, validation and test set) or with data from
another NLI task. In this study, we experiment with the same data of
the analysis, i.e., the entire MaClaim21 in the \cldom and \cl models
and Manifesto21 in the \dom and \none models. This means that each
sequence representation of the dataset is stacked into a matrix for
the computation of W and $\mu$.

\begin{table*}[]
\centering
\begin{tabular}{lcccccc}
\toprule
                                           &     & \multicolumn{2}{c}{\textbf{MaClaim21}}                                                                                                                                           & \cellcolor[HTML]{FFFFFF} & \multicolumn{2}{c}{\textbf{Manifesto21}}                                                                                                                                     \\ \cmidrule{3-4} \cmidrule{6-7}  
\multicolumn{1}{c}{\textbf{Model + postproc.}} & & \multicolumn{1}{c}{\textbf{\begin{tabular}[c]{@{}c@{}}\cldom\\ (++++)\end{tabular}}} & \multicolumn{1}{c}{\textbf{\begin{tabular}[c]{@{}c@{}}\cl\\ (+++)\end{tabular}}} & \cellcolor[HTML]{FFFFFF} & \multicolumn{1}{c}{\textbf{\begin{tabular}[c]{@{}c@{}}\dom\\ (++)\end{tabular}}} & \multicolumn{1}{c}{\textbf{\begin{tabular}[c]{@{}c@{}}\none\\ (+)\end{tabular}}} \\ 
\rowcolor[HTML]{FCFCD3} 
fasttext$_{avg}$                             & \cellcolor[HTML]{FFFFFF}  & 0.17                                                                                        & 0.30                                                                               & \cellcolor[HTML]{FFFFFF} & 0.27                                                                               & 0.28                                                                                    \\
\rowcolor[HTML]{FCFCD3} 
fasttext$_{avg}$+whiten                  &  \cellcolor[HTML]{FFFFFF}     & 0.54{$^{*}$}                                                                                       & 0.35                                                                               & \cellcolor[HTML]{FFFFFF} & 0.44{$^{*}$}                                                                               & 0.41                                                                                    \\
\rowcolor[HTML]{FAE1AE} 
BERT$_{german}$                           & \cellcolor[HTML]{FFFFFF}     & 0.12                                                                                        & 0.28                                                                               & \cellcolor[HTML]{FFFFFF} & 0.11                                                                               & 0.27                                                                                    \\
\rowcolor[HTML]{FAE1AE} 
BERT$_{german}$ +whiten                   &  \cellcolor[HTML]{FFFFFF}    & 0.37                                                                                        & 0.47{$^{*}$}                                                                               & \cellcolor[HTML]{FFFFFF} & 0.36                                                                               & 0.48{$^{*}$}                                                                                    \\
\rowcolor[HTML]{FCFCD3} 
RoBERTa$_{xml}$                        &   \cellcolor[HTML]{FFFFFF}      & 0.03                                                                                        & 0.35                                                                               & \cellcolor[HTML]{FFFFFF} & 0.08                                                                               & 0.33                                                                                    \\
\rowcolor[HTML]{FCFCD3} 
RoBERTa$_{xml}$+whiten                   &   \cellcolor[HTML]{FFFFFF}    & 0.39                                                                                        & 0.51{$^{*}$}                                                                               & \cellcolor[HTML]{FFFFFF} & 0.46{$^{*}$}                                                                               & 0.54{$^{*}$}                                                                                    \\
\rowcolor[HTML]{FAE1AE} 
SBERT                                      &  \cellcolor[HTML]{FFFFFF}   & 0.38                                                                                        & 0.47{$^{*}$}                                                                               & \cellcolor[HTML]{FFFFFF} & 0.31                                                                               & 0.47{$^{*}$}                                                                                    \\
\rowcolor[HTML]{FAE1AE} 
SBERT(whiten)                               & \cellcolor[HTML]{FFFFFF}   & \textbf{0.57{$^{*}$} }                                                                              & 0.50{$^{*}$}                                                                               & \cellcolor[HTML]{FFFFFF} & \textbf{0.53{$^{*}$} }                                                                     & 0.57{$^{*}$}                                                                                    \\
\rowcolor[HTML]{FCFCD3} 
SBERT$_{domain}$                           &\cellcolor[HTML]{FFFFFF}     & 0.22                                                                                        & 0.23                                                                               & \cellcolor[HTML]{FFFFFF} & 0.32                                                                               & 0.16                                                                                    \\
\rowcolor[HTML]{FCFCD3} 
SBERT$_{domain}$+whiten                    &  \cellcolor[HTML]{FFFFFF}   & 0.44{$^{*}$}                                                                                        & 0.45{$^{*}$}                                                                               & \cellcolor[HTML]{FFFFFF} & 0.41                                                                               & 0.52{$^{*}$}                                                                                    \\
\rowcolor[HTML]{FAE1AE} 
{\color[HTML]{333333} SBERT$_{party}$}     &  \cellcolor[HTML]{FFFFFF}   & {\color[HTML]{333333} 0.45}                                                                 & {\color[HTML]{333333} 0.13}                                                        & \cellcolor[HTML]{FFFFFF} & {\color[HTML]{333333} 0.32}                                                        & {\color[HTML]{333333} 0.16}                                                             \\
\rowcolor[HTML]{FAE1AE} 
{\color[HTML]{333333} SBERT$_{party}$+whiten}  &\cellcolor[HTML]{FFFFFF} & {\color[HTML]{333333} 0.53{$^{*}$} }                                                                & {\color[HTML]{333333} \textbf{0.70{$^{*}$} }}                                              & \cellcolor[HTML]{FFFFFF} & {\color[HTML]{333333} 0.50{$^{*}$} }                                                       & {\color[HTML]{333333} \textbf{0.69{$^{*}$} }}
\\ \bottomrule
\end{tabular}
\caption{Experimental results: Mantel's correlation between categorical and textual distance matrices. \textit{+whiten} means that the models have undergone whitening postprocessing. The + symbol indicates the level of informativeness from Table \ref{tab:eval-desc}. Highest correlation for each model in boldface. * p-value $<$ 0.05.}
\label{tab:results}
\end{table*}

\section{Experimental Setup}
\label{sec:experimental-setup}

\subsection{Datasets}

\paragraph{Fine tuning.} We use the German Manifesto data for 2013 and
2017 to fine-tune SBERT following Section~\ref{subsec:text-rep}.
There is a deliberate temporal gap between the fine tuning datasets
and the year of our ground truth, namely 2021, to ensure that the
model picks up generalizable differences between parties rather than
overfitting. However, we acknowledge the drawback that fine-tuning
does not receive any signal from newly emerged topics (e.g. Covid19)
and that party communication has not transformed drastically over the
last four years.

Appendix \ref{sec:appendix-a3} provides more details and statistics,
including evaluation on a 20\% held-out validation set, which shows
that fine-tuning improves both SBERT$_{party}$ and SBERT$_{domain}$
over plain SBERT, with SBERT$_{domain}$ gaining most.

\paragraph{Party representation.} To compute party similarities
following Section~\ref{sec:four-models-compute}, we use the 2021
manifestos, which arguably form the right textual basis to evaluate
against our Wahl-o-Mat ground truth for the 2021 German elections
(Section~\ref{sec:ground-truth:-wahl}). Recall that the Manifesto data
comes with annotated domains, but not with annotated claims. We
therefore applied an automatic claim classifier to identify claims
\cite{blokker-etal-2020-swimming}. We evaluated the results of the
classifier by calculating the precision on a subset of 324 manually
labeled claims from the 2021 manifestos and obtained a reasonable
precision of 75,6\%. More information about data and classifier can be
found in Appendix \ref{sec:appendix-c}.

This procedure results in two datasets for model training: Manifesto21
(with domain annotation) has 17,052 sentences; MaClaim21 (with domain
and claim annotation) consists of 9,814 claims. More details and
statistics are in Appendix \ref{sec:appendix-b}.

\subsection{Models}

In our empirical evaluation below, we vary the following three
parameters: (1), Embedding model and fine-tuning (SBERT plain
vs. SBERT$_{domain}$ vs. SBERT$_{party}$). (2), Party similarity
computation (\cldom vs. \cl vs. \dom vs. \none). (3), Postprocessing
(whitening vs. none). We consider all combinations of these
parameters.


\paragraph{Baselines}
We consider three baselines. The first and simplest one is a
pre-trained FastText model for German based on character $n$-gram
embeddings \citep{bojanowski2017enriching}. We compute sentence
representations by tokenizing the sentences based on the FastText tokenizer and averaging all FastText
token representations.\footnote{We evaluated both on the general
version of fasttext for German available on \url{fasttext.cc} and also
on a trained version with newspaper articles from TAZ for a more
domain specific model. Since both models obtained comparable
results, we report only results for the former.}

The other two baselines use transformer-driven (sub)word embeddings,
namely from BERT-German
\footnote{\url{https://huggingface.co/bert-base-german-cased}} and
multilingual RoBERTa-XLM
\footnote{\url{https://huggingface.co/xlm-roberta-base}}. We choose
the former because monolingual models often perform better than
multilingual ones and the latter because it is the student model with
which SBERT has been trained, which allows us to check how much better
SBERT can be in a text similarity task in the political domain. Again,
we feed each sentence to these models and compute the final
representations by averaging all token representations from the two
last layers of the model, a strong baseline for similarity tasks
\citep{li-etal-2020-sentence, Su2021WhiteningSR}.

\subsection{Evaluation}

To evaluate the pairwise party similarities computed by the models, we
turn them into distances and compare them against our ground truth
distance matrix (Section~\ref{sec:ground-truth:-wahl}) with the Mantel
test \citep{mantel1967detection}. This test is a variant of standard
correlation tests (such as Spearman's $rho$) which are not applicable to
distance matrices because they assume that the observations are
independent of one another. In our case, changing the position of one
value in the matrix would change the correlation between a pair or
parties. Having said that, the Mantel test addresses this problem by
calculating correlations on all permutations of the flattened distance
matrix. The two-tail hypothesis tests whether the correlation between
the ground truth matrix and the target distance matrix is
statistically significant or not. We use the nonparametric
version of the test since the party distances are not normally
distributed.

\section{Results and Discussion}
\label{sec:results-discussion}

Table \ref{tab:results} shows the quantitative results of our
experiments. We first discuss the effect of our various experimental
parameters.

\paragraph{Effect of postprocessing.} By comparing the upper and the
lower row in each colored block, we observe that the the whitening
transformation is beneficial in nearly all models, and where it is
not, the loss is minor.  On average, post-processed model embeddings
are 22 percentage points higher in the correlations, and consistently
obtain significant correlations with the ground truth. This suggests
that the benefit of enforcing isotropic distributions extends to the
domain and genre of political texts. Given the substantially higher performance of the models with the post-processing step, we focus on their results for the remainder of this
discussion.

\paragraph{Effect of embedding models and fine-tuning.} Comparing the
rows in the table, we observe that our two baseline models, BERT and
RoBERTa, show generally worse performance than even the non fine-tuned
SBERT. BERT is generally the worst performer among the three, despite
its monolinguality, which we interpret as evidence that the
architectures more geared towards similarity tasks have an advantage.
We take these results as validation of our choice of SBERT as
embedding model.

Interestingly, our simplest baseline, fasttext$_{avg}$, performs
better than most models in the most informative scenario (Mantel=0.54) and
relatively well with domain information (Mantel=0.44), but degrades when
less information is available. This suggests that FastText embeddings
are informative enough to support generalization from rich annotation,
but are not able to align semantically similar sentences well in a
less informative scenario such as in the twin matching approach.

Among the fine-tuned variants of SBERT, SBERT$_{domain}$ performs
surprisingly badly and is generally outperformed by vanilla
RoBERTa. This suggests that optimizing the model to pick up on domain
contrasts is distracting the model from capturing the dis(similarity)
between parties.

In contrast, SBERT$_{party}$ does very well, and competes with vanilla
SBERT for the best results. Indeed, SBERT wins in both setups that are
grouped by the domain category (\cldom and \dom), reaching 0.57 and
0.53, respectively. Conversely, SBERT$_{party}$ wins the two scenarios
without the grouping by domains (\cl's Mantel=0.70 and \none's Mantel=0.69), and achieves the
overall highest correlations here.

These results suggest that SBERT, without any fine-tuning, is
reasonably good at capturing the proximity between parties if more
information is provided: if we have both only claim structure and the
domain category then SBERT can be enough (Mantel=0.57). If there is
unstructured data, but there is still domain information, despite
having a drop in performance, it can still achieve a reasonable
correlation (Mantel=0.53).

SBERT$_{party}$, in contrast, performs better in the settings without
domain information, that is, when the party similarity is based on
twin sentence similarity (Section~\ref{sec:four-models-compute}). We
believe that this is the case because the sentence-level
fine-tuning of SBERT$_{party}$ is most directly carried forward into
the predictions of the model. In effect, therefore, fine-tuning SBERT
by contrasting the party difference is the best way to encode
fine-grained differences between parties' views and ideologies.

\begin{figure*}[]
\centering
  \subfloat[\cldom model (0.57)]{\includegraphics[width=0.5\columnwidth]{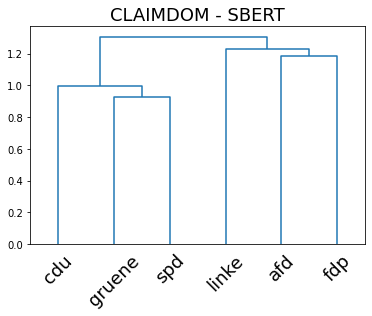}}
  \subfloat[\cl model (0.70)]{\includegraphics[width=0.5\columnwidth]{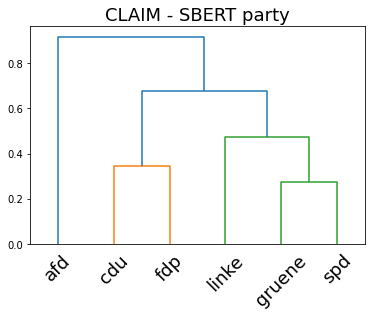}}
\subfloat[\dom model (0.53)]{\includegraphics[width=0.5\columnwidth]{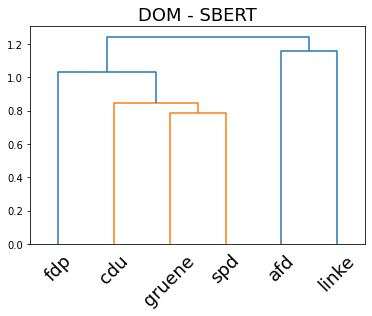}}
  \subfloat[\none model (0.69)]{\includegraphics[width=0.5\columnwidth]{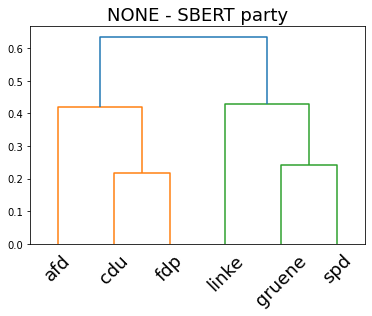}}
  \caption{Agglomerative clustering for the best model of each setting. Mantel correlation in parenthesis. Ground truth's comparison in Fig. \ref{fig:wom}b.}
  \label{fig:clusterings}
\end{figure*}

\paragraph{Analysis by agglomerative clustering.} To complement the
analysis by correlation coefficients in Table~\ref{tab:results}, we
compute agglomerative clusterings with average linkage for the best
models from Table~\ref{tab:results}. The results, shown in Figure
\ref{fig:clusterings}, show a good correspondence to the quantitative
results, thus lending support our use of the Mantel test.

Indeed, the two SBERT models in \ref{fig:clusterings}(a) and
\ref{fig:clusterings}(c), which reach moderate correlation
coefficients, disagree substantially with the ground truth clustering:
they group, for example, the far right AFD with the liberal FDP in
(a), and with the left wing Linke in (c). Also, the conservative CDU
is grouped with Grüne (greens) and social democratic SPD. In contrast,
the two SBERT$_{party}$ models in \ref{fig:clusterings}(b) and
\ref{fig:clusterings}(d) show a better match with the ground truth,
even though both group Grüne with SPD instead of Linke, and (b) has
AFD as an outlier altogether.

\paragraph{General outcome.} Probably the most striking outcome of our
experiment is that the best results -- both in terms of the
correlation coefficient and in terms of the clustering -- results from
models that use very little structured information (\cl, \none). The
difference among the two is small, and can be seen as a trade-off
between using a larger, more noisy dataset (all sentences:
Manifesto21) and a more focused dataset (just the claims: MaClaim21)
of about half the size.  These results confirm the idea that it is
possible to use natural language processing methods to identify the
dis(similarity) between party according to their policy positions with unstructured data.

We believe that this result is a combination of a good choice of
fine-tuning regimen -- providing the embeddings with a signal
concerning the contrast between parties -- with an appropriate way to
model similarity, with our twin matching approach which helps to match
the most relevant parts of the two manifestos to one another. These
two aspects reinforce each other, since a well fine-tuned model is
better able to push away dissimilar parties while bringing closer together similar ones.

\section{Conclusion}
\label{sec:conclusion}


In this paper, we have investigated to what degree text
representations can capture the proximity of parties and how to best
fine-tune representations for this task. Our results indicate that
aspects that have been proposed as important for this type of analysis
in political science, namely annotation of domains \citep{manifesto}
and claims \citep{koopmans1999political}, do not appear to matter
greatly for this task -- or at least, manual annotation can be
replaced by NLP tools: we have recognized claims with a classifier
\citep{blokker-etal-2020-swimming} and have proposed a weekly supervised
method, \enquote{twin matching}, to approximate domain-level
similarity computation. Indeed, one of our models that does not use
any manual annotation is among the top contenders. Of rather greater
importance for party similarity prediction, according to our findings,
is fine-tuning the text representations and postprocessing them.

This is good news for computational political science: the judicious
use of document structure appears able to help alleviate the effort of
having domain experts annotate large corpora. The two main limitations
of our current study relate to this outlook: (a) we only experimented
with a single language and ground truth -- future work should take
into account multiple languages and time periods, with a potential
long term goal of text-based models for party development
\cite{10.2307/24572675}; (b) we only scratched the surface of cues
available for fine-tuning. Future work could, for example, take into
account other aspects of parties such as ideological
position~\cite{glavas-etal-2017-unsupervised}, or reach beyond
manifestos to include information from other types of party
interactions \cite{10.2307/2111461}. In addition to that, work on
interpreting both the fine-tuned and vanilla SBERT models would be
interesting to better understand the predominant dimensions of the
sentence representations in the political domain.

\section*{Acknowledgments}
We acknowledge funding by Deutsche Forschungs- gemeinschaft (DFG) for project MARDY~2 (375875969) within the priority program RATIO.

\section*{Ethics Statement}
We believe that this study does not carry major ethical implications
in terms of data privacy or handling, given that our datasets are
based on publicly available party manifestos from the German elections
and from a public and freely accessible voting advice application
(Wahl-o-Mat). The annotators that provided us with a subset of labeled
claims to estimate the quality of the claim classifier were student
assistants from the university remunerated fairly according to their
working hours.

\bibliography{emnlp2022}

\begin{thebibliography}{28}
\expandafter\ifx\csname natexlab\endcsname\relax\def\natexlab#1{#1}\fi

\bibitem[{Blokker et~al.(2020)Blokker, Dayanik, Lapesa, and
  Pad{\'o}}]{blokker-etal-2020-swimming}
Nico Blokker, Erenay Dayanik, Gabriella Lapesa, and Sebastian Pad{\'o}. 2020.
\newblock \href {https://doi.org/10.18653/v1/2020.nlpcss-1.3} {Swimming with
  the tide? positional claim detection across political text types}.
\newblock In \emph{Proceedings of the Fourth Workshop on Natural Language
  Processing and Computational Social Science}, pages 24--34, Online.
  Association for Computational Linguistics.

\bibitem[{Bojanowski et~al.(2017)Bojanowski, Grave, Joulin, and
  Mikolov}]{bojanowski2017enriching}
Piotr Bojanowski, Edouard Grave, Armand Joulin, and Tomas Mikolov. 2017.
\newblock Enriching word vectors with subword information.
\newblock \emph{Transactions of the Association for Computational Linguistics},
  5:135--146.

\bibitem[{Brown et~al.(2020)Brown, Mann, Ryder, Subbiah, Kaplan, Dhariwal,
  Neelakantan, Shyam, Sastry, Askell, Agarwal, Herbert-Voss, Krueger, Henighan,
  Child, Ramesh, Ziegler, Wu, Winter, Hesse, Chen, Sigler, Litwin, Gray, Chess,
  Clark, Berner, McCandlish, Radford, Sutskever, and
  Amodei}]{NEURIPS2020_1457c0d6}
Tom Brown, Benjamin Mann, Nick Ryder, Melanie Subbiah, Jared~D Kaplan, Prafulla
  Dhariwal, Arvind Neelakantan, Pranav Shyam, Girish Sastry, Amanda Askell,
  Sandhini Agarwal, Ariel Herbert-Voss, Gretchen Krueger, Tom Henighan, Rewon
  Child, Aditya Ramesh, Daniel Ziegler, Jeffrey Wu, Clemens Winter, Chris
  Hesse, Mark Chen, Eric Sigler, Mateusz Litwin, Scott Gray, Benjamin Chess,
  Jack Clark, Christopher Berner, Sam McCandlish, Alec Radford, Ilya Sutskever,
  and Dario Amodei. 2020.
\newblock \href
  {https://proceedings.neurips.cc/paper/2020/file/1457c0d6bfcb4967418bfb8ac142f64a-Paper.pdf}
  {Language models are few-shot learners}.
\newblock In \emph{Advances in Neural Information Processing Systems},
  volume~33, pages 1877--1901.

\bibitem[{Budge(2003)}]{budge2003validating}
Ian Budge. 2003.
\newblock Validating the manifesto research group approach: theoretical
  assumptions and empirical confirmations.
\newblock In \emph{Estimating the policy position of political actors}, pages
  70--85. Routledge.

\bibitem[{Budge et~al.(2001)Budge, Klingemann, Volkens, Bara, and
  Tanenbaum}]{budge2001}
Ian Budge, Hans-Dieter Klingemann, Andrea Volkens, Judith Bara, and Eric
  Tanenbaum, editors. 2001.
\newblock \emph{Mapping {{Policy Preferences}}: {{Estimates}} for {{Parties}},
  {{Electors}}, and {{Governments}} 1945-1998}.
\newblock {Oxford University Press}, {Oxford, New York}.

\bibitem[{Burst et~al.(2021)Burst, Krause, Lehmann, Lewandowski, Matthieß,
  Merz, Regel, and Zehnter}]{manifesto}
Tobias Burst, Werner Krause, Pola Lehmann, Jirka Lewandowski, Theres Matthieß,
  Nicolas Merz, Sven Regel, and Lisa Zehnter. 2021.
\newblock Manifesto corpus. version: 2021.1.
\newblock \emph{Berlin: WZB Berlin Social Science Center.}

\bibitem[{Cer et~al.(2018)Cer, Yang, Kong, Hua, Limtiaco, St.~John, Constant,
  Guajardo-Cespedes, Yuan, Tar, Strope, and Kurzweil}]{cer-etal-2018-universal}
Daniel Cer, Yinfei Yang, Sheng-yi Kong, Nan Hua, Nicole Limtiaco, Rhomni
  St.~John, Noah Constant, Mario Guajardo-Cespedes, Steve Yuan, Chris Tar,
  Brian Strope, and Ray Kurzweil. 2018.
\newblock \href {https://doi.org/10.18653/v1/D18-2029} {Universal sentence
  encoder for {E}nglish}.
\newblock In \emph{Proceedings of the 2018 Conference on Empirical Methods in
  Natural Language Processing: System Demonstrations}, pages 169--174,
  Brussels, Belgium. Association for Computational Linguistics.

\bibitem[{Devlin et~al.(2019)Devlin, Chang, Lee, and
  Toutanova}]{devlin-etal-2019-bert}
Jacob Devlin, Ming-Wei Chang, Kenton Lee, and Kristina Toutanova. 2019.
\newblock \href {https://doi.org/10.18653/v1/N19-1423} {{BERT}: Pre-training of
  deep bidirectional transformers for language understanding}.
\newblock In \emph{Proceedings of the 2019 Conference of the North {A}merican
  Chapter of the Association for Computational Linguistics: Human Language
  Technologies, Volume 1 (Long and Short Papers)}, pages 4171--4186,
  Minneapolis, Minnesota. Association for Computational Linguistics.

\bibitem[{Dor et~al.(2018)Dor, Mass, Halfon, Venezian, Shnayderman, Aharonov,
  and Slonim}]{ein-dor-etal-2018-learning}
Liat~Ein Dor, Yosi Mass, Alon Halfon, Elad Venezian, Ilya Shnayderman, Ranit
  Aharonov, and Noam Slonim. 2018.
\newblock \href {https://doi.org/10.18653/v1/P18-2009} {Learning thematic
  similarity metric from article sections using triplet networks}.
\newblock In \emph{Proceedings of the 56th Annual Meeting of the Association
  for Computational Linguistics (Volume 2: Short Papers)}, pages 49--54,
  Melbourne, Australia. Association for Computational Linguistics.

\bibitem[{Druckman et~al.(2005)Druckman, Martin, and
  Thies}]{druckman2005influence}
James~N Druckman, Lanny~W Martin, and Michael~F Thies. 2005.
\newblock Influence without confidence: Upper chambers and government
  formation.
\newblock \emph{Legislative Studies Quarterly}, 30(4):529--548.

\bibitem[{Ethayarajh(2019)}]{ethayarajh-2019-contextual}
Kawin Ethayarajh. 2019.
\newblock \href {https://doi.org/10.18653/v1/D19-1006} {How contextual are
  contextualized word representations? {C}omparing the geometry of {BERT},
  {ELM}o, and {GPT}-2 embeddings}.
\newblock In \emph{Proceedings of the 2019 Conference on Empirical Methods in
  Natural Language Processing and the 9th International Joint Conference on
  Natural Language Processing (EMNLP-IJCNLP)}, pages 55--65, Hong Kong, China.
  Association for Computational Linguistics.

\bibitem[{Gao et~al.(2019)Gao, He, Tan, Qin, Wang, and Liu}]{gao-2019}
Jun Gao, Di~He, Xu~Tan, Tao Qin, Liwei Wang, and Tie{-}Yan Liu. 2019.
\newblock \href {https://openreview.net/forum?id=SkEYojRqtm} {Representation
  degeneration problem in training natural language generation models}.
\newblock In \emph{7th International Conference on Learning Representations,
  {ICLR} 2019, New Orleans, LA, USA, May 6-9, 2019}. OpenReview.net.

\bibitem[{Gao et~al.(2021)Gao, Yao, and Chen}]{gao-etal-2021-simcse}
Tianyu Gao, Xingcheng Yao, and Danqi Chen. 2021.
\newblock \href {https://doi.org/10.18653/v1/2021.emnlp-main.552} {{S}im{CSE}:
  Simple contrastive learning of sentence embeddings}.
\newblock In \emph{Proceedings of the 2021 Conference on Empirical Methods in
  Natural Language Processing}, pages 6894--6910, Online and Punta Cana,
  Dominican Republic. Association for Computational Linguistics.

\bibitem[{Glava{\v{s}} et~al.(2017)Glava{\v{s}}, Nanni, and
  Ponzetto}]{glavas-etal-2017-unsupervised}
Goran Glava{\v{s}}, Federico Nanni, and Simone~Paolo Ponzetto. 2017.
\newblock \href {https://aclanthology.org/E17-2109} {Unsupervised cross-lingual
  scaling of political texts}.
\newblock In \emph{Proceedings of the 15th Conference of the {E}uropean Chapter
  of the Association for Computational Linguistics: Volume 2, Short Papers},
  pages 688--693, Valencia, Spain. Association for Computational Linguistics.

\bibitem[{Koopmans and Statham(1999)}]{koopmans1999political}
Ruud Koopmans and Paul Statham. 1999.
\newblock Political claims analysis: Integrating protest event and political
  discourse approaches.
\newblock \emph{Mobilization: an international quarterly}, 4(2):203--221.

\bibitem[{König et~al.(2013)König, Marbach, and
  Osnabrügge}]{10.2307/24572675}
Thomas König, Moritz Marbach, and Moritz Osnabrügge. 2013.
\newblock \href {http://www.jstor.org/stable/24572675} {Estimating party
  positions across countries and time—a dynamic latent variable model for
  manifesto data}.
\newblock \emph{Political Analysis}, 21(4):468--491.

\bibitem[{Lapesa et~al.(2020)Lapesa, Blessing, Blokker, Dayan{\i}k, Haunss,
  Kuhn, and Pad{\'o}}]{lapesa2020debatenet}
Gabriella Lapesa, Andr{\'e} Blessing, Nico Blokker, Erenay Dayan{\i}k,
  Sebastian Haunss, Jonas Kuhn, and Sebastian Pad{\'o}. 2020.
\newblock {DEbateNet}-mig15: tracing the 2015 immigration debate in germany
  over time.
\newblock In \emph{Proceedings of The 12th Language Resources and Evaluation
  Conference}, pages 919--927.

\bibitem[{Laver et~al.(2003)Laver, Benoit, and Garry}]{laver2003extracting}
Michael Laver, Kenneth Benoit, and John Garry. 2003.
\newblock Extracting policy positions from political texts using words as data.
\newblock \emph{American political science review}, 97(2):311--331.

\bibitem[{Li et~al.(2020)Li, Zhou, He, Wang, Yang, and
  Li}]{li-etal-2020-sentence}
Bohan Li, Hao Zhou, Junxian He, Mingxuan Wang, Yiming Yang, and Lei Li. 2020.
\newblock \href {https://doi.org/10.18653/v1/2020.emnlp-main.733} {On the
  sentence embeddings from pre-trained language models}.
\newblock In \emph{Proceedings of the 2020 Conference on Empirical Methods in
  Natural Language Processing (EMNLP)}, pages 9119--9130, Online. Association
  for Computational Linguistics.

\bibitem[{Liu et~al.(2020)Liu, Ott, Goyal, Du, Joshi, Chen, Levy, Lewis,
  Zettlemoyer, and Stoyanov}]{liu2020roberta}
Yinhan Liu, Myle Ott, Naman Goyal, Jingfei Du, Mandar Joshi, Danqi Chen, Omer
  Levy, Mike Lewis, Luke Zettlemoyer, and Veselin Stoyanov. 2020.
\newblock \href {https://doi.org/10.48550/arXiv.1907.11692} {Ro{BERT}a: A
  robustly optimized {BERT} pretraining approach}.
\newblock \emph{ArXiv}, abs/1907.11692.

\bibitem[{Mantel(1967)}]{mantel1967detection}
Nathan Mantel. 1967.
\newblock The detection of disease clustering and a generalized regression
  approach.
\newblock \emph{Cancer research}, 27(2):209--220.

\bibitem[{McGregor(2013)}]{mcgregor2013measuring}
R~Michael McGregor. 2013.
\newblock Measuring “correct voting” using comparative manifestos project
  data.
\newblock \emph{Journal of Elections, Public Opinion and Parties}, 23(1):1--26.

\bibitem[{Nanni et~al.(2022)Nanni, Glava{\v{s}}, Rehbein, Ponzetto, and
  Stuckenschmidt}]{nanni2022political}
Federico Nanni, Goran Glava{\v{s}}, Ines Rehbein, Simone~Paolo Ponzetto, and
  Heiner Stuckenschmidt. 2022.
\newblock Political text scaling meets computational semantics.
\newblock \emph{ACM/IMS Transactions on Data Science (TDS)}, 2(4):1--27.

\bibitem[{Reimers and Gurevych(2019)}]{reimers2019sentencebert}
Nils Reimers and Iryna Gurevych. 2019.
\newblock \href
  {http://dblp.uni-trier.de/db/conf/emnlp/emnlp2019-1.html#ReimersG19}
  {Sentence-bert: Sentence embeddings using siamese bert-networks}.
\newblock In \emph{Proceedings of EMNLP/IJCNLP}, pages 3980--3990. Association
  for Computational Linguistics.

\bibitem[{Slapin and Proksch(2008)}]{slapin2008scaling}
Jonathan~B Slapin and Sven-Oliver Proksch. 2008.
\newblock A scaling model for estimating time-series party positions from
  texts.
\newblock \emph{American Journal of Political Science}, 52(3):705--722.

\bibitem[{Strom(1990)}]{10.2307/2111461}
Kaare Strom. 1990.
\newblock \href {http://www.jstor.org/stable/2111461} {A behavioral theory of
  competitive political parties}.
\newblock \emph{American Journal of Political Science}, 34(2):565--598.

\bibitem[{Su et~al.(2021)Su, Cao, Liu, and Ou}]{Su2021WhiteningSR}
Jianlin Su, Jiarun Cao, Weijie Liu, and Yangyiwen Ou. 2021.
\newblock \href {https://arxiv.org/abs/2103.15316} {Whitening sentence
  representations for better semantics and faster retrieval}.
\newblock \emph{ArXiv}, abs/2103.15316.

\bibitem[{Wagner and Ruusuvirta(2012)}]{wagner2012}
Markus Wagner and Outi Ruusuvirta. 2012.
\newblock \href {https://doi.org/10.1057/ap.2011.29} {Matching voters to
  parties: {{Voting}} advice applications and models of party choice}.
\newblock \emph{Acta Politica}, 47(4):400--422.

\end{thebibliography}
\bibliographystyle{acl_natbib}

\setcounter{table}{0}
\onecolumn
\appendix

\section{Appendix}

\subsection{Fine-tuning data}
\label{sec:appendix}

\begin{table}[H]
\centering
\begin{tabular}{lc}
\textbf{Party} & \textbf{Num. inst.} \\ \hline
Grüne                               & 5913                                              \\ 
Die Linke                            & 4243                                              \\ 
Social Democratic Party of Germany (SPD)                                  & 3566                                              \\ 
Free Democratic Party (FDP)                                  & 3149                                              \\ 
Christian Democratic Union (CDU)                                  & 2569                                              \\ 
Alternative for Germany (AfD)                                  & 770                                               \\ 
\end{tabular}
\caption{Number of instances in the train set of the fine-tuning of SBERT$_{party}$. Data from the 2013 and 2017 manifestos.}
\label{tab:by-party}
\end{table}

\begin{table}[H]
\centering
\begin{tabular}{lc}
\textbf{Domain name}        & \textbf{Num. inst} \\ \hline
Welfare and Quality of Life & 7078                  \\ 
Economy                     & 6330                  \\ 
Fabric of Society           & 2586                  \\ 
Freedom and Democracy       & 2395                  \\ 
External Relations          & 2306                  \\ 
Social Groups               & 2144                  \\ 
Political System            & 1682                  \\ 
\end{tabular}
\caption{Number of instances in the train set of the fine-tuning of SBERT$_{domain}$. Data from the 2013 and 2017 manifestos.  More information about the categories can be found on \url{https://manifesto-project.wzb.eu/coding_schemes/mp_v5}}
\label{tab:by-domain}
\end{table}

\begin{table}[H]
\centering
\begin{tabular}{llp{0.4\linewidth}l}
\textbf{Party} & \textbf{Year} & \textbf{Sentence}                                                                                                                                                                                       & \textbf{Domain}           \\ \hline
AfD            & 2017          & This oligarchy holds the levers of state power, political education and informational and media influence over the population.                                       & Political System            \\
CDU            & 2017          & We have set ourselves an ambitious goal: We want full employment for  all of Germany by 2025 at the latest.                                                          & Social Groups               \\
FDP            & 2013          & We want to continue to give people the freedom to pursue their ideas - creating growth, progress and prosperity for all.                                              & Freedom and Democracy       \\
Grüne          & 2013          & We want to make a change today to move towards an economy that benefits  everyone, not just a few.                                                                   & Welfare and Quality of Life \\
Die Linke      & 2013          & But the populations and workers of these countries have common interests: the fight against wage depression, recession and mass unemployment.              & Economy                     \\
SPD            & 2017          & This includes ensuring that social cohesion in our country becomes stronger again  and that decent dealings with one another are not lost to political radicalization. & Fabric of Society          
\end{tabular}
\caption{Examples from the training dataset with their corresponding domain names translated from German.}
\label{tab:ex-manifesto}
\end{table}

\subsection{S-BERT training parameters}
\label{sec:appendix-a2}

\begin{itemize}
    \setlength\itemsep{0em}
    \item Pre-trained model: paraphrase-multilingual-mpnet-base-v2
    \item Maximum sequence length: 128
    \item Train batch size: 16
    \item Number of training epochs: 5
    \item Learning rate: 2e-5
    \item Warm up steps: 100
\end{itemize}

\subsection{Fine-tuning evaluation}
\label{sec:appendix-a3}

\begin{table}[H]
\centering
\begin{tabular}{lll}
\textbf{Model} & \textbf{f1} & \textbf{SBERT (f1)} \\ \hline
SBERT$_{domain}$      & 71,39\%      & 66,66\%                     \\ 
SBERT$_{party}$       & 68,79\%      & 66,66\%                     \\ 
\end{tabular}
\caption{Comparison of the f1 scores between the non-fine-tuned  and fine-tuned SBERT models on the held out validation set. }
\label{tab:res-finetuning}
\end{table}

\section{Appendix}
\label{sec:appendix-b}

\subsection{Data for the evaluation}

\begin{table}[H]
\centering
\begin{tabular}{lc}
\textbf{Party} & \textbf{Num. claims} \\ \hline
Die Linke      & 2770                 \\
Gruene         & 2380                 \\
CDU            & 1685                 \\
FDP            & 1388                 \\
SPD            & 952                  \\
AfD            & 638                 
\end{tabular}
\caption{Number of claims per party in MaClaim21.}
\label{tab:macov-numclaims}
\end{table}

\begin{table}[H]
\centering
\begin{tabular}{lc}
\textbf{Party} & \textbf{Num. sentences} \\ \hline
Die Linke      & 4850                 \\
Gruene         & 3947                 \\
CDU            & 2775                 \\
FDP            & 2239                 \\
SPD            & 1665                 \\
AfD            & 1574                
\end{tabular}
\caption{Number of sentences per party in Manifesto21. }
\label{tab:maauto-numclaims}
\end{table}

\section{Appendix}
\subsection{Claim identifier}
\label{sec:appendix-c}

The claim identifier was trained on annotated data from the DebateNet dataset \citep{lapesa2020debatenet}. The annotations are based on news articles from the German newspaper TAZ regarding the migration in the domestic scenario. Sentences that contain a claim are considered as positive and sentences without any claims are negative. It has been verified that the claim identifier trained on DebateNet can transfer reasonably well to the party manifestos \cite{blokker-etal-2020-swimming} with an averaged f1 score of 82\% across the election campaigns of 2013 and 2017.  More information regarding the training process: 
\begin{itemize}
    \setlength\itemsep{0em}
    \item Number of training instances: 13,283
    \item Number of validation instances: 1,477
    \item Number of testing instances: 1,641
    \item Maximum sequence length: 128
    \item Train batch size: 32
    \item Number of training epochs: 5
    \item Learning rate: 3e-5
\end{itemize}

\subsection{Evaluation on 2021 party manifestos}

Expert annotators from the political science faculty annotated 324 unique political claims from six major German parties competing in the federal election of 2021. Annotations of claims followed a fine-grained hierarchical ontology (\textit{codebook}) yielding 75 unique sub-categories that are divided into eight major categories. While the latter broadly corresponds to relevant policy fields, such as \enquote*{health}, \enquote*{economy and finance}, or \enquote*{education}, the former specifies the concrete policy measure to be taken, for instance, \enquote*{mandatory vaccination}, \enquote*{raise taxes}, \enquote*{expansion of education and care services}. We do not provide the inter-annotator agreement because annotators worked closely together in this task. However, we verified the quality of the dataset by having a third annotator gold standardizing the dataset.

The classifier detected 245 out of the 324 annotated claims, reaching a reasonable precision of 75,6\%. In total, the classifier predicted 9,814 claims out of 17,052 sentences.

\end{document}